\pdfoutput=1
\documentclass[11pt]{article}

\usepackage[final]{acl}
\usepackage{times}
\usepackage{latexsym}
\usepackage[T1]{fontenc}
\usepackage[utf8]{inputenc}
\usepackage{microtype}
\usepackage{inconsolata}
\usepackage{graphicx}
\usepackage{booktabs}
\usepackage{array}
\usepackage{enumerate}
\usepackage{enumitem}
\usepackage[most]{tcolorbox}
\usepackage{forest}
\usepackage{xcolor}
\usepackage{xspace}
\newcommand{\ra}{$\rightarrow$\xspace}
\usepackage{tipa}
\usepackage{linguex}
\title{LLMs as a synthesis\\ between symbolic and distributed approaches to language}

\author{Gemma Boleda\\
  Universitat Pompeu Fabra / ICREA\\
  \texttt{gemma.boleda@upf.edu}\\}

\begin{document}

\maketitle

\begin{abstract}
    Since the middle of the 20th century, a fierce battle is being fought between symbolic and distributed approaches to language and cognition.
  The success of deep learning models, and LLMs in particular, has been alternatively taken as showing that the distributed camp has won, or dismissed as an irrelevant engineering development.
  In this position paper, I argue that deep learning models for language
  actually represent a synthesis between the two traditions.
  This is because 1)~deep learning architectures allow for both distributed/continuous/fuzzy and
  symbolic/discrete/categorical-like representations and processing; 2)~models trained on language make use of this flexibility.
  In particular, I review recent research in interpretability that showcases how a substantial part of morphosyntactic knowledge is encoded in a near-discrete fashion in LLMs.
  This line of research suggests that different behaviors arise in an emergent fashion, and
  models flexibly alternate between the two modes (and everything in between) as needed.
  This is possibly one of the main reasons for their wild success; and it makes them particularly interesting for the study of language.
  Is it time for peace?

\end{abstract}

\section{Introduction}

\begin{figure*}[htb]
  \centering
  \begin{tcolorbox}[title=SYMBOLIC]

    \begin{minipage}{0.25\textwidth}
      \centering
      \begin{tabular}{lll}
        S & \ra & NP VP\\
        VP & \ra & V NP PP\\
        NP & \ra & Det N | John | Mary\\
        PP & \ra & P NP\\
        Det & \ra & a\\
        N  & \ra & drink\\
        P  & \ra & to\\
        V & \ra & gave\\
      \end{tabular}
      
      \label{fig:grammar}
    \end{minipage} \hfill
    \begin{minipage}{0.65\textwidth}
      \centering
      \begin{forest}
        [S
        [NP [John]]
        [VP 
        [V [gave]]
        [NP 
        [Det [a]]
        [N [drink]]
        ]
        [PP 
        [P [to]]
        [NP [Mary]]
        ]
        ]
        ]
      \end{forest}
    \end{minipage}
  \end{tcolorbox}

  \begin{tcolorbox}[title=DISTRIBUTED AND (NEAR-)SYMBOLIC]
    
    \begin{minipage}{0.25\textwidth}
      \centering
      \includegraphics[width=\textwidth]{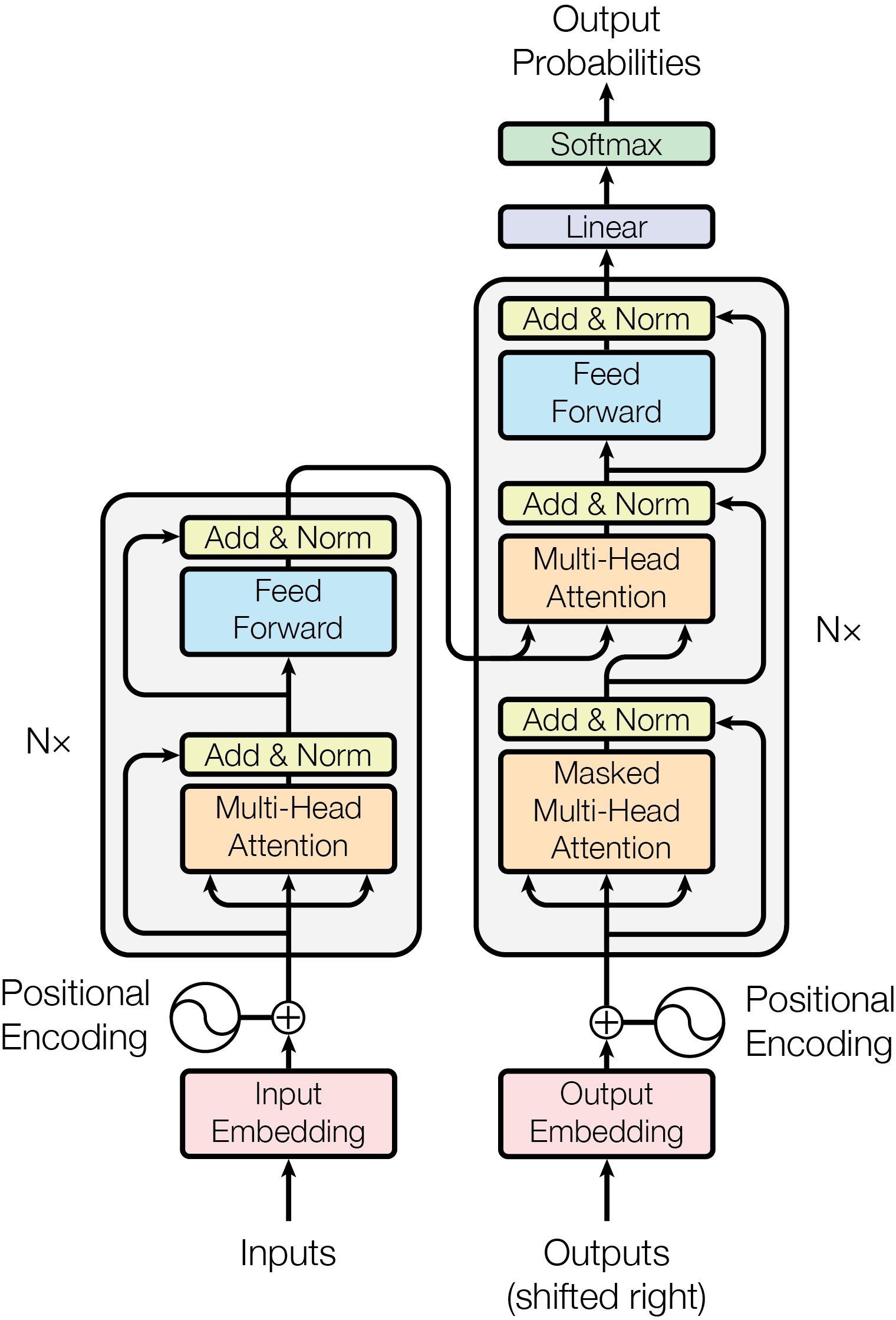}
      
    \end{minipage} \hfill
    \begin{minipage}{0.65\textwidth}
      \centering
      \includegraphics[width=\textwidth]{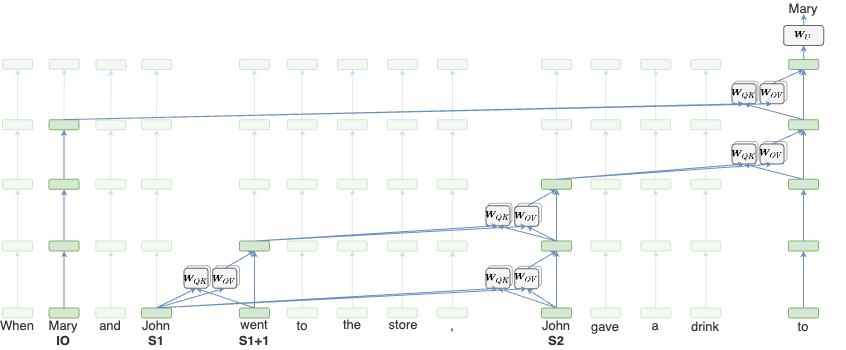}
      
    \end{minipage}
  \end{tcolorbox}
  
  \caption{Schematic illustration of the contrast between symbolic formalisms and deep learning. Top: context-free grammar and parse tree for the sentence ``John gave a drink to Mary". Bottom: transformer architecture and circuit for the fragment ``When Mary and John went to the store, John gave a drink to", with prediction ``Mary'' (adapted from \citet{vaswani2017attention} and \citet{ferrando2024primerinnerworkingstransformerbased}, with permission). In the circuit, the representations are continuous (vectors), but the different components function together in an interpretable algorithm, with attention heads carrying operations such as copying (see text for details).}
   \label{fig:main-fig}
\end{figure*}

Since the middle of the 20th century, a fierce battle is being fought between two antagonistic approaches to language and cognition. Although the specifics vary, they can be broadly characterized as follows.
Symbolic approaches typically work with discrete, interpretable categories (like ``noun'', ``verb'' for parts of speech) and discrete, interpretable rules to combine them (such as those of formal grammars).%
\footnote{In early work in NLP, these approaches were paired with top-down processing of linguistic data, through rule-based systems defined by hand. In later work, the processing part has instead been data-driven: data is manually annotated according to a given representation system, and a processing algorithm is induced from the data via machine learning. The latter includes modern neural networks trained for, e.g., dependency parsing. This means that the border between symbolic and distributed approaches is, quite fittingly with this papper, blurry. Relatedly, within formal linguistics different approaches have started softening the discreteness of both categories and rules~\cite[see e.g.][for a comprehensive discussion of probabilistic approaches to semantics and pragmatics]{Erk2022-prob}. Still, even in these cases the most common approach is to add probabilities or constraints to symbolically-defined rules and categories.}
Distributed approaches instead couple uninterpretable high-dimensional continuous representations, such as vectors, with continuous functions to combine them, such as those defined in the different components of a Transformer.%
\footnote{\label{fn:symb-dist} I use ``symbolic'' and ``distributed'' as umbrella terms, with related notions being discrete/categorical/localist for the former, and continuous/fuzzy/sub-symbolic for the latter. The different terms touch on different properties that for the purposes of this paper can be lumped together; I will make nuances explicit when needed.}

The debate between the two approaches has taken different forms in different fields: classicism vs.\ connectionism in in cognitive science \cite{sep-connectionism}, symbolic / rule-based vs.\ data-driven / Machine Learning-based approaches in AI~\cite{Russell2020}, formalism / generativism vs.\ functionalism / cognitivism in linguistics \cite{harris2021linguistics}.%
\footnote{Functionalists do not use distributed representations, but the issues underlying the divide between formalists and functionalists are very related to the general debate, as will become clear during the article, so I am including functionalism in the distributed camp. Also note that the respective positions are rooted in the philosophical traditions of rationalism and empiricism~\cite{sep-rationalism-empiricism}.}
The crux of the debate is that, across all these fields, some researchers focus on the rule-like behavior of language and cognition and others on its slippery nature.

The advent of deep learning has added fuel to the scientific fire.
In some circles, the success of deep learning models has been alternatively taken as showing that the distributed camp has won, or dismissed as an irrelevant engineering development.
A prime example is Steve Piantadosi's 2024 provocatively titled article ``Modern language models refute Chomsky's approach to language'' \cite{piantadosi2024} and the answers it has received, some as heated as ``Modern Language Models Refute Nothing'' \cite[]{rawski-baumont23}; another is the also provocative squib by the late Felix Hill, titled ``Why transformers are \textit{obviously} good models of language'' \cite[emphasis in the original]{hill2024transformersobviouslygoodmodels}. 
I believe that these maximalist positions are sterile.

In this position paper, I join more constructive voices exploring what deep learning models might contribute to linguistic theory~\cite{manning15,warstadt2022artificial,futrell2025linguisticslearnedstopworrying}.
In particular, I propose that LLMs are actually a \textbf{synthesis} between symbolic and distributed approaches to language (Figure~\ref{fig:main-fig}).%
\footnote{I center the discussion on LLMs as the most widely adopted type of model, but I will also touch on other kinds of models, such as neural machine translation models.}

This view rests on two theses.
The first is that the debate exists precisely because language is \textbf{both} symbolic (or discrete) and distributed (or fuzzy)---and everything in between.
Indeed, the sustained debate between the two approaches suggests that neither is able to capture language on its own~\cite{boleda-herbelot-2016-formal}; and it is necessary to move towards integrated models that capture the full spectrum of language.
The second thesis is that modern LLMs are one such kind of model, because they support both distributed and (near-)symbolic representations and processing.
In my view, this is one of the main reasons for their amazing success at language.

Now, the synthesis view of LLMs may come as a surprise, since neural networks undoubtedly fall in the distributed camp; however, what is often overlooked in the debate is the fact that neural networks do have the potential for near-symbolic representations and processing \cite[see Section~\ref{sec:llms-both} for discussion]{Smolensky2006}.
Crucially, however, this potential still leaves open what newer-generation neural network models will do with it in practice.
My argument is based on recent results in the interpretability literature which suggest that, when deep learning models are exposed to language data and are asked to do a predictive task like language modeling, they develop near-discrete representations and quasi-symbolic processes in addition to distributed ones.

Figure~\ref{fig:main-fig} schematically illustrates the contrast between symbolic formalisms and deep learning architectures as I see it, with the example of syntax: while symbolic formalisms are discrete, neural networks afford both distributed and near-discrete representations and processes.

The contributions of the paper are as follows:
\vspace{-0.2cm}
\begin{itemize}
  \itemsep -0.08cm
\item summarizing for the CL/NLP community the ways in which language, as a phenomenon, exhibits both regularity and messiness (Thesis~1; Section~\ref{sec:lang-both});
\item appraising recent interpretability work that suggests that LLMs deploy near-symbolic representations and processes in addition to distributed ones (Thesis 2; Section~\ref{sec:llms-both});
\item explaining how this situates LLMs in a debate that has permeated the study of cognition since the 1950s (remainder of the paper).
\end{itemize}

\section{Language is both regular and messy}
\label{sec:lang-both}

Regarding Thesis 1, let's start by exemplifying clear cases of regularity.
In morphosyntax, for instance, it is common to posit that words belong to different parts of speech (such as noun or verb).%
\footnote{The exact shape that this takes depends on the theory, with some theories placing more strength in the grammar and others on the lexicon~\cite[see][for discussion]{Borer2017}. The difficulties discussed in this section surface in both kinds of theories, though in different ways.}
Languages mark morphosyntax formally, and the combinatorics of linguistic units are governed by morphosyntactic properties. For instance, the English suffix \textit{-ed} marks tense, and only verbs inflect for tense (\textit{follow/followed}, but \textit{before/*befored}).
Similarly, syntactic phrases can stand in different syntactic relations (such as subject, object, or indirect object), which can also be formally marked.
For instance, the indirect object in English is marked by the preposition \textit{to}, as in example~\ref{ex15}.
In many languages, different units standing in a given syntactic relation display agreement~\cite{wechsler2003many}.
For instance, in English, subjects and verbs agree in number; in example~\ref{ex2}, the singular subject (\textit{A student}) must appear with a singular verb (\textit{is}).
In Spanish, there is gender and number agreement also within the noun phrase: in example~\ref{ex:partes}, the highlighted suffix \textit{-a} on the determiner and adjective mark feminine gender, in agreement with the noun's lexical gender.

\ex. \label{ex15} John gave a drink to/*for Mary

\ex. \label{ex2} A student is/*are crossing the street

\ex. \label{ex:partes}
    \gll L\textbf{\underline{a}}s partes interesad\textbf{\underline{a}}s\\
     the.F.PL party.PL interested.F.PL\\
    \glt `The interested parties.'

    In the syntax-semantics interface, a classic phenomenon is anaphora, with syntactic constraints determining the shape of anaphoric pronouns: for instance, in~\ref{ex3}, the pronoun \textit{him} cannot refer to Mark~\cite{Chomsky1981}.
    As an example from compositional semantics, it is well known that adding negation in a sentential context reverses polarity  \cite[][see example~\ref{ex6}]{zeijlstra2007negation}.

\ex. \label{ex3} Mark$_i$ combs himself$_i$/*him$_i$

\ex. \label{ex6} I will/will not come to lunch

All of these phenomena are categorical or discrete, in that there is no ``in between'' state: verbs inflect for tense, prepositions don't; \textit{is} is right and \textit{are} wrong in the context of singular subjects; \textit{not} is a like a binary switch for polarity; etc.
Moreover, in all of them, we find a systematic relationship between form and function, or grammar and meaning, such as \textit{-ed} marking past tense.

This kind of data is what spurred symbolic approaches, where discrete symbols are combined via discrete rules relying on formal features, across domains as different as phonology~\cite{chomsky1968sound,PrinceSmolensky1993}, syntax~\cite{Chomsky1957,KaplanBresnan1982,Langacker1987,gazdar1985generalized,pollard1994head}, semantics~\cite{montague74:EFL,kamp-reyle93,partee+90,pustejovsky95}, and pragmatics~\cite{sperber-wilson95,roberts2012information,webber2016formal}.%
\footnote{I will mainly discuss morphosyntax and semantics, for two reasons: Because these domains are representative of the issues that underlie the debate between symbolic and distributed approaches, and because most of the work on LLM interpretability is in these domains (and the latter is the literature that provides the empirical basis for the synthesis view).
However, in Appendix~\ref{app:other-ling-domains} I also briefly discuss phonology and morphology. Also note that I will also mostly use English examples for space reasons. Nothing in my argument hinges on this choice.}

However, one needs only scratch the surface for regularity to break down.
The border between parts of speech is notoriously fuzzy~\cite{Croft2001,Evans2009};
there is no universal agreed upon set of syntactic relations~\cite{Napoli1993,van2005exploring}; negation is far from being a binary switch in many contexts (e.g., \textit{not unhappy} does not mean \textit{happy}), and is hugely complex from a semantic point of view~\cite{zeijlstra2007negation}; and even agreement can break down~\cite{wechsler2003many}.

In my view, messiness comes from two main sources.
First, \textbf{fuzzy borders between categories} like those of parts of speech are pervasive across linguistic domains~\cite{Croft2001,dowty1991thematic,haspelmath2007pre}.%
\footnote{To the point that scholars have often questioned the existence of many categories~\cite[see e.g.][for parts of speech]{Croft2001}. There is an important theoretical distinction between ascertaining the existence of a given theoretical construct (e.g.\ in the mind/brain) and gauging its usefulness as a scientific tool. The discussion in this paper is aimed at the former; but the data I discuss cannot distinguish between the two levels.}
Continuing with the example of parts of speech (see Appendix~\ref{app:other-ling-domains} for other examples), in many Indo-European languages there is much fuzziness between adjectives and nouns, nouns and verbs, and adjectives and verbs; so much so that, when manually POS-tagging a corpus, a common recourse is to allow for multiple tags~\cite{marcus-etal-1993-building}.
An example is shown in~\ref{ex:fright1}, where \textit{frightened} could be either a verbal participle, interpreted as in~\ref{ex:frightA}, or an adjective denoting an emotional state, as in~\ref{ex:frightB} (analogous to \textit{sad}, \textit{happy}).%
\footnote{While context often disambiguates, discussing the manual annotation of the Penn Treebank, \citet[p.\ 316]{marcus-etal-1993-building} note that ``even given explicit criteria for assigning POS tags to potentially ambiguous words, [sometimes] the word's part of speech \textit{simply cannot be decided}'' (my emphasis).}

\ex. \label{ex:fright1} The frightened child
    \a. The child who was frightened by something/someone \label{ex:frightA}
    \b. The child feeling fright \label{ex:frightB}

The other way in which languages resist symbolic treatment is by breaking down the systematic \textbf{relationship between form and function}.
Clear examples are irregular or semi-regular morphological forms, arising from historical processes~\cite{matthews1991morphology}.
For instance, in many English verbs the past tense is not marked by \textit{-ed}, but by an irregular form (\textit{went}, \textit{was}) or a semi-regular pattern (e.g.\ the so-called ablaut pattern in forms such as \textit{sang}, \textit{drank}, \textit{began}).

While these are purely formal irregularities, most form-function mismatches actually result from an interaction between grammar and meaning.
Example~\ref{ex5} showcases agreement \textit{ad sensum}: In sharp contrast to \ref{ex2} above, here a plural verb is allowed despite the fact that the subject is headed by a singular noun.
Agreement \textit{ad sensum} usually happens with singular head nouns that denote sets or pluralities, such as \textit{group} ---i.e., cases where there is a mismatch between grammatical and semantic features.

\ex. \label{ex5} A group of students from New Zealand is/are crossing the street

This kind of semantic leakage into syntax poses a hurdle to symbolic approaches based solely on formal features.
Within a symbolic framework, it is still possible to add constraints that take into account the semantics of the head noun in computing agreement, for instance by adding a \textsc{denotes-plurality} feature to the representation of the noun.
And, while approaches to this phenomenon in formal linguistics are highly sophisticated, they involve integrating semantics along these lines~\cite{wechsler2003many}.
This is an apparently easy fix, which however opens a path fraught with difficulties, as encoding conceptual aspects of meaning in a discrete way is arguably unfeasible (see below).

In formal linguistics, the difficulty has been handled by strictly distinguishing semantic features that are grammatically relevant from those that ``merely'' constitute world knowledge, and circumscribing the empirical scope of linguistic theory to the former~\cite{Jackendoff1990,levin93}.
However, as pointed out in functional approaches like cognitive linguistics, interactions between grammar and conceptual aspects of meaning are pervasive in language; and there is no clear dividing line between semantic properties that are relevant vs.\ irrelevant for grammar~\cite{Langacker1987,Fillmore1988,goldberg95}.
Thus, the strict division betweeen linguistic-semantics and other-semantics is questionable. Moreover, it narrows the empirical scope of linguisticy theory, delegating many language phenomena to other disciplines.
Conversely, the risk in functional approaches, given the difficulties involved in encoding the relevant factors, is to forfeit the predictive power of linguistic theories, thus staying at a descriptive level.

And so we enter the ultimate messy place in language: conceptual aspects of meaning.
The clearest example is word meaning, which is notoriously fuzzy, vague, and slippery~\cite{wittgenstein53,kilgarriff97,boleda2020distributional}.
For instance, in contrast to cases like \ref{ex15} and \ref{ex2} above, the similarities and differences between \textit{fast} and \textit{swift} are subtle, and there is no hard and fast rule to determine when to use one and when to use the other.
Trying to delimit a word's meaning is similarly challenging;
\citet{wittgenstein53} famously discussed the case of \textit{game},
concluding that there are no necessary and sufficient conditions determining what counts as a game, and all we can ask for is some kind of ``family resemblance''.
This is why dealing with lexical semantics in terms of discrete features, such as \textsc{denotes-plurality}, is fraught with difficulties.

For these reasons, if a fully symbolic approach to parts of speech is problematic, a fully symbolic approach to lexical semantics has been argued to be ultimately unfeasible; very prominently in our community \cite{boleda2020distributional}, but also in other traditions from philosophy~\cite{wittgenstein53,gardenfors2014geometry} to lexicography~\cite{kilgarriff97,hanks2000word}.
And, indeed, despite monumental efforts building lexical resources like WordNet, or developing systems for tasks like Word Sense Disambiguation, our community could not model word meaning at scale until distributed methods came along.
That being said, even within this messy domain we still find categorical distinctions.
For instance, while different word senses are often impossible to delineate precisely~\cite{kilgarriff97}, in some cases the distinction is very clear, e.g.\ the \textsc{animal} and \textsc{computer device} senses of \textit{mouse}; and some concepts are crisper than others (e.g.\ \textsc{five} vs. \textsc{fast}). 

And other aspects of semantics are largely discrete and symbolic, notably reference~\cite{frege1892}.
We use language to talk about the world and, from a linguistic point of view, there is nothing fuzzy in the distinction between, say, two people with the same name. Thus, whether \textit{Elizabeth Blackburn won the Nobel prize} is true will depend on which Elizabeth Blackburn we're talking about in the given context.%
\footnote{As of 2025, according to the internet there are at least two Elizabeth Blackburns: a Nobel laureate and a judge in Florida.}
To sum up, this overview suggests that language is indeed both discrete and fuzzy; and that there is no neat discrete/fuzzy divide, nor any area of language that is completely discrete or completely fuzzy.
At the same time, there \textit{are} clearly areas that are more discrete (such as morphosyntax) and areas that are fuzzier (such as word meaning).

Crucially, no scholar questions any of the empirical data I have discussed so far; what changes is the way they are appraised.
Some traditions focus on the regularities and consider the rest as either special cases or phenomena outside the purview of linguistic theory; whereas others sustain that the ubiquity of these ``special cases'' makes the regularities an epiphenomenon at best~\cite{weissweiler2025linguistic}.
These are conscious choices that are based on carefully considered theoretical positions.
The clearest example of this dichotomy is the aforementioned generative vs.\ cognitive divide, with generative linguistics tending towards the former~\cite[among many others]{Chomsky1957,KaplanBresnan1982,montague74:EFL} and cognitive linguistics towards the latter~\cite[again among many others]{Langacker1987,Fillmore1988,goldberg95}.
My tenet here is that \textbf{both properties are fundamental}, and we cannot reduce language to one or the other.
Therefore, we need models that natively encompass both regularity and messiness.

\section{LLMs and regularity}
\label{sec:llms-both}

To my knowledge, it has not been contested that neural networks in general, and LLMs in particular, can do fuzzy processing of the sort required for e.g.\ lexical semantics.
Therefore, here I will place my emphasis on regularity (Appendix~\ref{app:llms-messy} briefly discusses non-symbolic and non-compositional processing in LLMs).

\begin{figure}[tb]
  \includegraphics[width=\columnwidth]{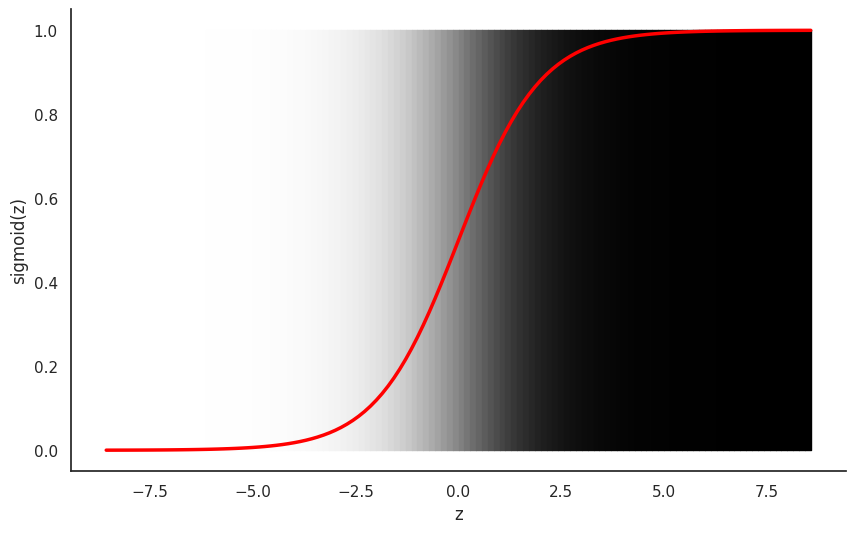}
  \caption{Non-linear functions such as the sigmoid provide the potential for both continuous and near-discrete behavior.}
  \label{fig:sigmoid}
\end{figure}

Indeed, the main criticism of neural networks has historically been their inadequacy in handling rule-like linguistic behavior (see e.g.\ \citealp{manning15} and \citealp{pinker1988language}).
However, the distributed camp has long argued that the architecture of neural networks affords symbolic-like processing \cite{rumelhart-mclelland86tenses,minsky1988perceptrons,Smolensky2006,Smolensky2012}.
Recall from above that distributed approaches couple high-dimensional representations with continuous functions to combine them; importantly, however, some of these functions are nonlinear, and this is what gives neural networks the potential for rule-like behavior~\cite{minsky1988perceptrons}.
Take the sigmoid as an example (Figure~\ref{fig:sigmoid}): when its input falls near 0, the value passed on will be continuous; but when its input is larger or smaller, it will be quasi-binary.
This allows networks to learn to combine their inputs in a way that leverages non-linearities to build more or less distributed representations and processing, as needed.

If we put this potential together with the properties of language discussed in the previous section, we can expect LLMs to exploit this potential when trained on language.
And this is indeed what recent literature on interpretability suggests.

However, what counts as near-symbolic behavior in the context of neural networks?
While this is a very difficult notion to pin down, in this paper I count as near-symbolic the existence of small sub-units of the network that are causally involved in encoding or processing a single linguistic construct.%
\footnote{This definition does not imply that this sub-unit need be the only one involved in the relevant behavior; see Section~\ref{sec:discussion} for discussion.}
This sub-unit can be at different levels of description, from single neurons to larger components like attention heads.

It has been known for close to a decade that neural LMs encode non-trivial knowledge of syntax, including its hierarchical nature~\cite{linzen-etal-2016-assessing,gulordava-etal-2018-colorless,futrell-etal-2019-neural,rogers2021primer}. However, most earlier work used techniques such as probing, which could show THAT they encode syntactic knowledge, but not HOW.
Newer methods in interpretability \cite[see][for a survey]{ferrando2024primerinnerworkingstransformerbased} focus on precisely this question, and it is these methods that have provided the clearest evidence for near-discreteness in some aspects of linguistic processing in deep learning models.%
\footnote{The vast majority of results in this literature concerns English; in what follows, I'll refer to results for English.}
Most studies focus on morphosyntactic properties or syntactic relations.

\paragraph{Neurons.} Several studies have identified neurons that selectively respond to morphosyntactic properties such as part of speech, number, and tense~\cite{bau2019identifying,durrani+2023,gurnee2023findingneuronshaystackcase,gurnee2024universal}. For instance, \citet{durrani+2023} find neurons sensitive to part of speech in three multi-lingual LLMs (BERT, RoBERTa, and XLNet), such as neuron 624 in layer 9 of RoBERTa responding to verbs in the simple past tense and neuron 750 in layer 2 to verbs in the present continuous tense.
Moreover, some morphosyntactic neurons are ``universal'' \cite{gurnee2024universal}, in the sense that they can be found across different instantiations of the same auto-regressive LLM.
This suggests that language data provide a strong pressure for neurons encoding morphosyntactic properties to arise.

Other studies look at the effects of specific neurons on the output~\cite{geva-etal-2022-lm,geva-etal-2022-transformer,ferrando-etal-2023-explaining}.
\citet{geva-etal-2022-transformer} identified neurons that drastically promote the prediction of tokens with specific features, some of which are morphosyntactic in nature; for instance, neuron 1900 in layer 8 of GPT2 increased the probability of WH words (e.g.\ \textit{which}, \textit{where}, \textit{who}) and neuron 3025 in layer 6 of WikiLM the probability of adverbs (e.g.\ \textit{largely}, \textit{rapidly}, \textit{effectively}).
Relatedly, \citet{ferrando-etal-2023-explaining} identified a small set of neurons that are functionally active in making grammatically correct predictions (for instance in subject-verb agreement) in models of the GPT2, OPT, and BLOOM families.

My favorite example regarding neurons is \citet{bau2019identifying}, who analyzed neurons associated to morphosyntactic properties in a neural Machine Translation model from the pre-transformer era.
Altering the values of these neurons changed the morphosyntactic properties of the translations.
For example, altering the activation of a single encoder neuron changed the translation of the whole phrase \textit{The interested parties} into Spanish, switching its gender from feminine to masculine (cf.~\ref{ex:partes2}, with the highlighted feminine \textit{-a} vs.\ masculine \textit{-o} gender suffixes).
Remarkably, both translations are correct, but they convey different meanings: the feminine noun \textit{parte} is a general equivalent of \textit{party}, and the masculine \textit{partido} in this context implies specifically a political party.

\ex. \label{ex:partes2} The interested parties\\
  \textit{Original}: L\textbf{\underline{a}}s partes interesad\textbf{\underline{a}}s\\
  \textit{Modified}: L\textbf{\underline{o}}s partid{\textbf{\underline{o}}}s interesad\textbf{\underline{o}}s\\

\begin{figure}[tb]
  \centering

  (a)
  
  \includegraphics[width=.75\columnwidth]{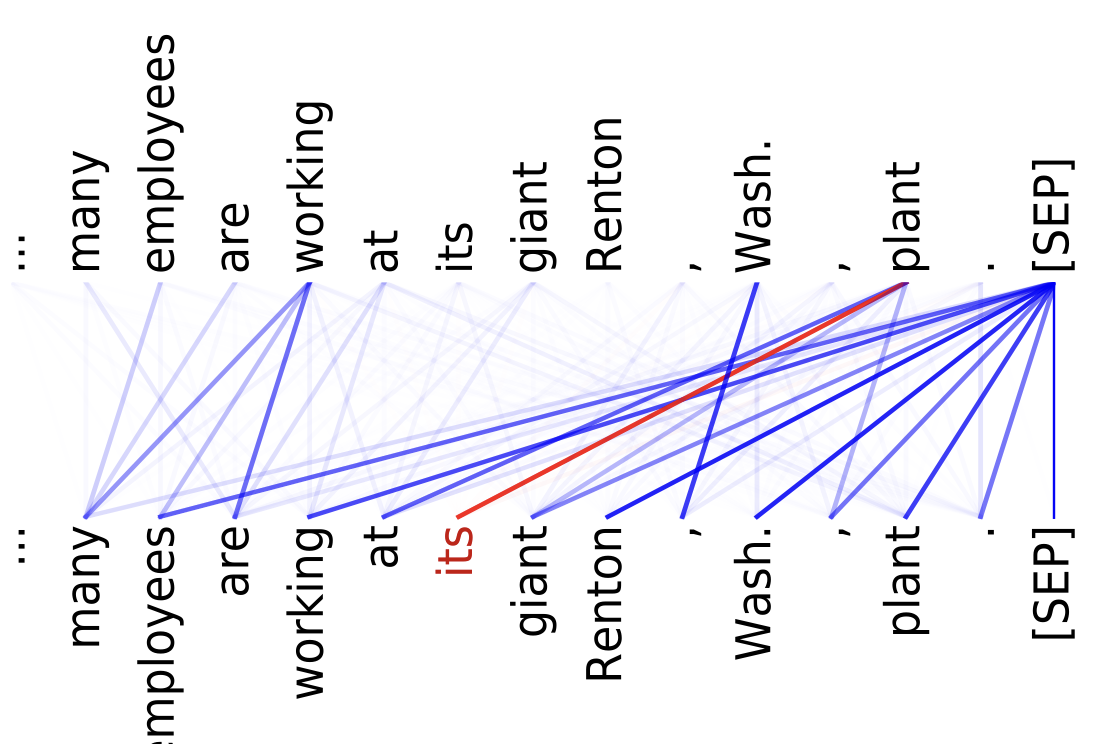}

  (b)
  \vspace{.1cm}

  \includegraphics[width=.75\columnwidth]{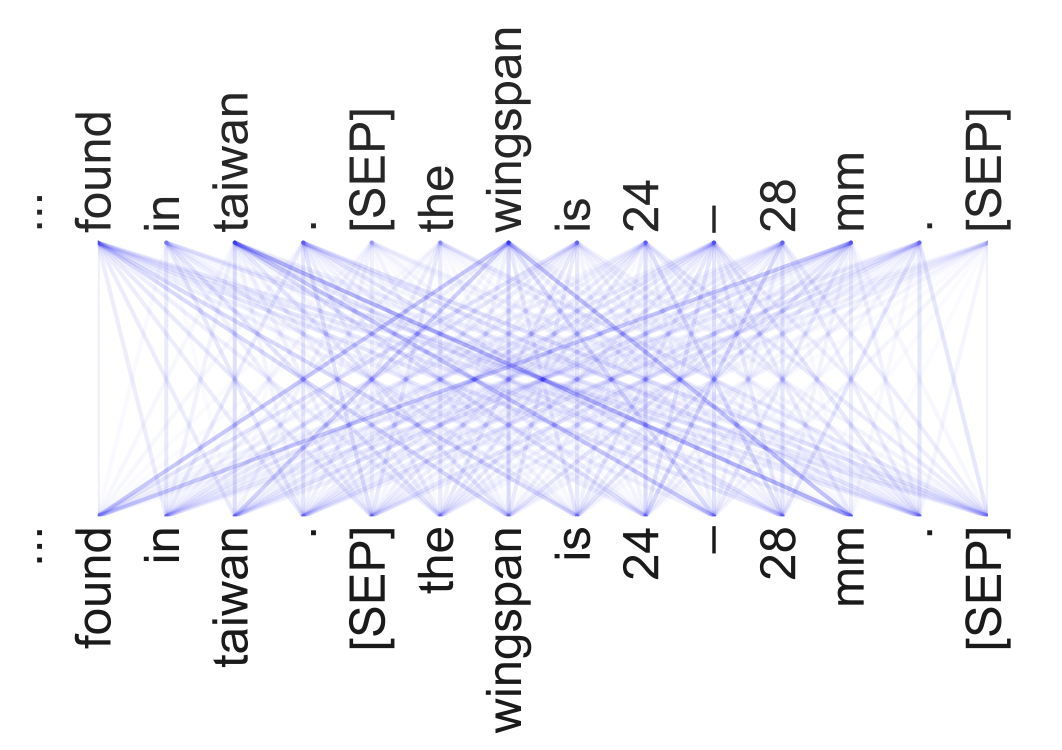}
  
  \caption{Near-discrete and continuous attention heads in BERT (adapted from \citet{clark-etal-2019-bert}, CC-BY license; line thickness is proportional to amount of attention). (a) Head 7 in layer 6 tracks dependencies between possessive determiners and their head nouns dependency in a near-discrete fashion: the determiner \textit{its}, highlighted in red, sharply attends to its head noun \textit{plant}. (Note that most tokens have near-discrete attention to the [SEP] token. \citet{clark-etal-2019-bert} interpreted this as a no-op signal.) (b) Head 1 in layer 1 instead presents a broad attention pattern with no clear interpretation.}
  \label{fig:det-noun}
\end{figure}

\paragraph{Attention heads.} Attention heads with specialized syntactic functions have also been widely found in LLMs and neural MT models~\cite{raganato-tiedemann-2018-analysis,clark-etal-2019-bert,htut2019attentionheadsberttrack,voita-etal-2019-analyzing,krzyzanowski+24}.
Figure~\ref{fig:det-noun}(a) shows the activations of BERT's head 7 in layer 6 for the sentence \textit{many employees are working at its giant Renton, Wash., plant}. This head specializes in the possessive construction; in the example, the possessive determiner (\textit{its}) sharply attends to its head noun (\textit{plant}), in a dependency relation that has 5 intervening tokens in the surface structure.
Other heads highlighted in this literature correspond to a wide range of syntactic relations such as subject, object, prepositional complement, adjectival modifier, or adverbial modifier.
Note that all heads are near-discrete; Figure~\ref{fig:det-noun}(b) depicts a head with a broad attention pattern.
The existence of these broad heads again suggests the need for distributed processing of other properties, which are however more difficult to interpret.

\paragraph{Circuits.} In recent years, more evidence has emerged around the notion of ``circuit'', or subgraphs within neural networks \cite{cammarata2020thread:}.%
\footnote{More specifically a circuit is ``A subgraph of a neural network. Nodes correspond to neurons or directions (linear combinations of neurons). Two nodes have an edge between them if they are in adjacent layers. The edges have weights which are the weights between those neurons [...]''  \cite{olah2020zoom}.}
A particularly relevant example for us is \citet{wang2023interpretability}, 
which describes in detail a circuit in GPT2-small governing the prediction of the indirect object of a sentence.
Figure~\ref{fig:main-fig} (bottom right) contains a schematic depiction of the circuit for the sentence \textit{When John and Mary went to the store, John gave a drink to \_\_}, where the LLM predicts \textit{Mary}.
This interpretable circuit corresponds to an algorithm that identifies the names in the sentence (in the example, \textit{John} and \textit{Mary}), removes the names that appear in the second sentence (\textit{John}), and outputs the remaining name (\textit{Mary}).
The model does this through different attention heads with specialized functions.
\citet{merullo2024circuit} further provide evidence that this circuit is robust (they identify the same circuit in a larger GPT2 model) and generalizes: some of its individual components are reused for a task that is different both semantically and syntactically (it involves the generation of a word denoting the color of an object described among other objects in the preceding context).
This suggests that the uncovered circuit is at a quite high level of abstraction in terms of linguistic knowledge.
\citet{ferrando-costa-jussa-2024-similarity} contribute further evidence of abstract generalization in circuits. They show that one and the same circuit is responsible for solving subject-verb agreement in English and Spanish in the multi-lingual LLM Gemma 2B.

\begin{figure}[tb]
  \centering
  \includegraphics[width=.75\columnwidth]{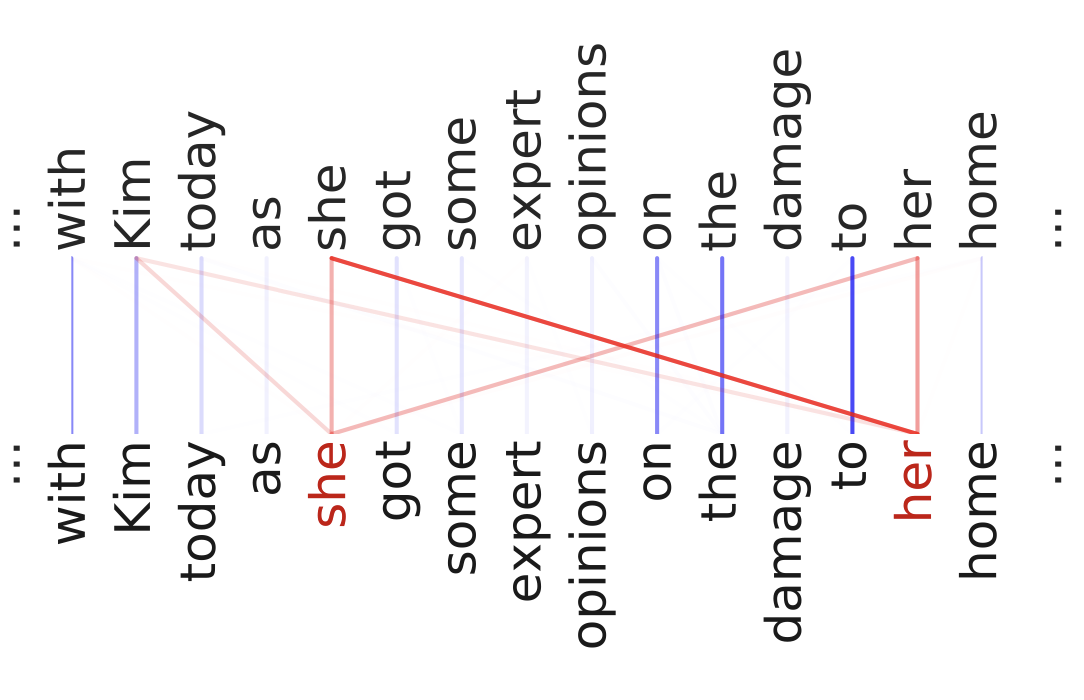}
  
  \caption{BERT's attention head tracks co-reference dependencies (head 5 in layer 4); adapted from \citet{clark-etal-2019-bert}. The anaphoric pronoun ``her'' sharply attends to antecedent ``she''.}
  \label{fig:coref}
\end{figure}

To sum up, the interpretability literature provides evidence for near-symbolic morphosyntactic processing in different sub-units of LLMs (neurons, attention heads, circuits).
Much less attention has been devoted to other domains, such as compositional semantics and the syntax-semantic interface, but the existing evidence points in the same direction.
For instance, BERT has attention heads specializing in co-reference, in which anaphoric mentions sharply attend to their antecedent~\cite[][see Figure~\ref{fig:coref}]{clark-etal-2019-bert}; and one of the aforementioned ``universal neurons'' in \citet{gurnee2024universal} selectively responds to negation.%
\footnote{The emergence of discrete behavior, and prominently circuits, has been related to what has been called ``grokking'' \cite{grokking}, that is, the sudden appearance of generalization capabilities in symbolic tasks. See e.g.\ \citet{nanda2023progress} and \citet{varma2023explaininggrokkingcircuitefficiency} for discussion. Here I focus on symbolic behavior in linguistic representations and processing, but of course its emergence in learning is an exciting topic for further study.}

\section{Discussion}
\label{sec:discussion}

The preceding section has explored the near-symbolic encoding and processing of linguistic information within LLMs.
However, as mentioned in the introduction, deep learning models can flexibly switch between discrete and distributed modes, and everything in between.
In this, they are very different from formalisms and representations used in theoretical linguistics.

Indeed, as emphasized throughout this paper, while representations in theoretical linguistics are discrete, in LLMs they are at most \textit{near}-discrete.
Moreover, there is wide variation in the degree of discreteness (!) exhibited with respect to different phenomena, or even within a phenomenon.
For instance, in the work cited above, \citet{durrani+2023} found drastically fewer neurons responding to the POS of function words (like determiners or numerals) than to the POS of content words (like nouns and verbs).
They conjectured that the representation of POS in the networks may be more distributed in the latter than in the former case.
Similarly, \citet{bau2019identifying} find that gender and number are represented in a more distributed fashion than tense in the NMT model they analyze.

Another crucial difference with classical formalisms in linguistics is the fact that in neural networks there is a high degree of redundancy \cite{durrani+2023}.
For instance, when \citet{wang2023interpretability} ablated some of the heads that they identified in the indirect object circuit explained above, they found that the circuit still worked to some extent. They subsequently went on to identify back-up heads that replaced the role of the initially identified heads.
Redundancy is a well-known property of neural networks, and one crucial for their functioning, as it allows for graceful as opposed to catastrophic degradation in behavior \cite{lecun+89}.

The flip side of redundancy is polysemanticity, that is, the fact that units respond to different properties~\cite{rumelhart1986parallel}.
For instance, in many (but not all) cases a neuron that responds to, say, tense, will also respond to some other unrelated property.
In a fine-grained analysis of GPT2-small attention heads including manual annotation, \citet{krzyzanowski+24} found that around 90\% are polysemantic.
There are advantages to polysemanticity, such as the fact that it allows networks to represent more features than they have dimensions~\cite[][call this ``superposition'']{elhage2022superposition}.

If we put the two features together (redundancy and polysemanticity), we see that each feature is represented across different individual neurons and neurons are responsible for different features.
By definition, this is what makes a representation distributed~\cite{Hinton1986}.
So why am I arguing that LLMs are a synthesis between continuous and discrete approaches?
Because, as a matter of fact, even if they could represent and process everything in a distributed fashion, they do not. They learn to process some aspects of language in a near-symbolic manner, to the point that specific interpretable algorithms can be reverse-engineered~\cite{ferrando-costa-jussa-2024-similarity}.
The 90\% figure just mentioned, from \citet{krzyzanowski+24}, implies that 10\% of the attention heads analyzed are monosemantic ---when they would not need to be, and in fact \textit{poly}semanticity has advantages, as mentioned above.
Similarly, most of the ``universal neurons'' identified by \citet{gurnee2024universal} are monosemantic, and they have clear functional roles in circuits, such as deactivating attention heads.
This stands in stark contrast to, for instance, the much more distributed representation of words in static or contextualized word embeddings.
And, indeed, the evidence for near-discrete behavior overwhelmingly comes from domains where symbolic formalisms have been the most successful, such as grammar.

In the context of this paper, it is important to distinguisth between \textit{symbolic} and \textit{interpretable}.
This paper's metareviewer remarked that ``the paper proposes symbolic representations that lack the properties that make symbolic representations appealing to most researchers, namely that they are interpretable by humans. The authors need to either make the point that the semi-discrete representations and rules in LLMs are actually interpretable in the way that traditional symbols and rules are, or make a case for symbolic representation separate from interpretability.''
My view falls squarely on the latter side. I do not think we can expect LLMs to ever amount to a complete symbolic framework; nor that they \textit{should}, because in my view language is not completely symbolic either. Therefore, the implication of my paper in this regard is that, if we aim at obtaining more understandable and transparent model architectures, we cannot simply aim at reducing LLMs to symbolic systems.%
\footnote{We can of course still extract symbolic knowledge for specific phenomena, and this can be very useful, the same way that symbolic frameworks are very useful in many domains.}
Instead, we need to devise
methods that embrace the full symbolic-to-distributed spectrum. Since (as far as I can tell) these methods do not exist yet, the only roadmap I can offer at present is to point out that we need to find new roads.

Relatedly, in using models to elucidate how language works, we should remember that the ultimate testing ground for theories of language and cognition is the brain.
Recent work suggests that there may be analogies between LLM and brain encoding of language~\cite{Tuckute2024}.
However, while research in neuroscience has yielded quite robust results on the different brain regions where language encoding takes place, and some of their roles, it has made much less progress on the
properties of linguistic representations and the computations that are carried out during processing~\cite{Tuckute2024}, namely, on the topic of this paper.
This is another very exciting avenue for further work.

\section{Conclusion}

I started this piece by pointing out that a fierce battle is being fought, since the second half of the 20th century, between symbolic and distributed approaches to language and cognition.
And I actually find it worrying that much of this discussion is being led by scholars outside the CL/NLP community.
Since we know the most about LLMs, we should participate in ascertaining what they tell us (and what they can't tell us yet) about how language works. One of the motivations of my paper is precisely to foster this kind of debate within our community.

The view I have put forth in this paper is that LLMs are a synthesis between the two approaches; they allow us to integrate regularity and messiness into a single modeling tool, thus overcoming the difficulties faced by symbolic-only or distributed-only systems.
Importantly, more and less distributed representations and processing arise in an emergent fashion; LLMs \textbf{learn} to behave in a quasi-symbolic fashion at times, in a highly fuzzy and distributed fashion at others, because that allows them to perform better at linguistic tasks, that is, they do so responding to pressures from language data.

So, may it be time for peace? The research I have surveyed has only scratched the surface, and we need everyone on board to continue to make progress in our collective understanding of how language works.
In particular, we need methods that go beyond specific, cherry-picked phenomena and allow a systematic exploration of the models~\cite[is a relevant step in this direction]{ferrando-voita-2024-information}; a better systematization of the empirical landscape to be explored~\cite[e.g.,][propose to broaden the benchmarks by which we evaluate the linguistic abilities of LLMs]{weissweiler2025linguistic}; and a stronger engagement with theory when evaluating the implications of deep learning models of language.

\section*{Limitations}

I am aware that my definition of what counts as near-symbolic in LLMs is, ironically, fuzzy.
I think that, given the present state of the art (mechanistic interpretation of deep learning models is still in its infancy), the best I can do is offer an initial definition and many examples of the kind of behavior that I think provides support for my position.
Delineating it more precisely is a pressing need for the future.

As a reviewer pointed out, the interpretability literature ``has not yet shown that these localized mechanisms truly function as symbolic components in a larger sense, nor has it demonstrated how they can scale to capture the kinds of generalizable rules or logical inferences that symbolic systems have historically handled''. While this is falls outside the scope of the article, it is worth stating that there is much less research in these kinds of phenomena.
Interestingly, however, there is at least some tentative evidence for parts of models working as symbolic components.
The study of \citet{merullo2024circuit} discussed above suggests systematic component reuse; similarly, \citet{lindsey2025biology} show how Claude 3.5 Haiku performs multi-hop reasoning re-using components across a range of phenomena. One of the cases they discuss is ``addition circuitry [that] generalizes between very different contexts''. This circuitry selectively activates in contexts where it’s useful to perform implicit addition, and includes mechanisms to ``represent and store intermediate computations for later use''.
These are just preliminary findings, and this is certainly another area where much more research is needed, as is research on the interplay between linguistic and reasoning abilities more generally.

\section*{Acknowledgments}
Thank you to Marco Baroni, Rafael Gutiérrez, Louise McNally and the members of the COLT research group for precious feedback.
This research is an output of grant PID2020-112602GB-I00/MICIN/AEI/10.13039/501100011033, funded by the Ministerio de Ciencia e Innovación and the Agencia Estatal de Investigación (Spain).

\bibliography{anthology,custom,../../gemma}
\appendix

\section{Regularities and messiness across domains}
\label{app:other-ling-domains}

As mentioned in the main section of the paper, we find regularity and messiness throughout the different aspects and domains of language.
Syntax and semantics were discussed in the main text; here we add some brief notes about phonology and morphology.

In phonetics and phonology, ``there is accumulating evidence that the categorical and continuous aspects of speech are deeply intertwined''~\cite{Roessig2019}.
A basic aspect is a language's phonological inventory, i.e.\ the set of phonemes that constitute it. Phonemes are ``the smallest contrastive sound unit[s] in a language that can distinguish meaning''; for instance, in Catalan, /l/ and /r/ are phonemes (e.g.\ \textit{cara} means `face', \textit{cala} `small bay'), but in Japanese they are allophones, that is, different phonetic realizations or pronunciations of the same phoneme.
There is ample evidence of categorical conceptualization of continuous speech by speakers into phonemes, but also equally ample evidence for challenges to a purely symbolic treatment of phonology, analogous to those discussed in the main text for parts of speech~\cite{Pierrehumbert16}.
Similarly, different phonemes combine according to phonotactic rules or constraints (e.g.\ in English, but not Catalan, the sequence of phonemes /sp/ can begin a syllable, e.g.\ in the word \textit{spa}), and those face difficulties analogous to those of syntax, with semantics leaking into phonotactics~\cite[e.g.][]{Baroni2001-ri}.%
\footnote{In Catalan, \textit{spa} is pronounced /\textipa{@s'pa}/.}

As for morphology, a basic notion such as that of word is as common as it is controversial and challenging to delimit~\cite{haspelmath2011indeterminacy}.
Similarly, inflection and derivation are considered fundamentally different kinds of processes, but their border is again fuzzy \cite{copot_idiosyncratic_2022,haley_corpus-based_2024}.
Moreover, we find that derivational morphology, like inflectional morphology, presents pervasive regularities together with irregularities and semi-regularities \cite{matthews1991morphology}.
For instance, the English suffix \textit{-ion} selects for verbal roots and produces nouns referring to actions, processes, or results (create/creation, operate/operation, donate/donation).
However, this pattern has many exceptions due to historical borrowing, primarily from Latin and French.
Many verbs and their corresponding nouns were borrowed into English as separate words, preserving irregularities from the original language (\textit{destroy/*destroyion/destruction}, from Latin \textit{destructio} \textit{admit/*admition/admission}, from Latin \textit{admissio}).
Remember that we discussed an analogous case with irregular verbs in English in the main text.
Furthermore, derivational morphology also displays the semi-regular match between form and meaning that we found in morphosyntax~\cite{lieber2004morphology,boleda2020distributional}.

\section{Non-symbolic and non-compositional linguistic processing in LLMs}
\label{app:llms-messy}

In the main text I have taken for granted that LLMs can do non-symbolic and non-compositional linguistic processing. Here I am presenting evidence for the latter, for completeness.
The realm with the richest evidence of non-symbolic processing is that of conceptual aspects of meaning, which as discussed in the paper defy symbolic treatment. I discuss two representative examples, lexical semantics and sentential semantics, but the evidence is vast.

As for lexical semantics, recall that, to account for a word's meaning and usage, symbolic methods like those in traditional Word Sense Disambiguation define a set of senses for each word and assign each use of a word in context to one of the senses. This has long been known to be problematic, as many sense boundaries are blurry and word usages can be more and less similar to each other~\cite{kilgarriff97}. LLMs provide instead graded representations for words in context, and this has been linked to their leap in success. Among the many papers about this, let me point to two specific analyses using BERT: \citet{gari-soler-etal-2019-word} show that BERT-estimated similarity between word usages corresponds to human similarity scores; and \citet{gari-soler-apidianaki-2021-lets} show that ``BERT representations offer good estimates of the partitionability of words into senses'', that is, to how easy or difficult it is to define different senses for a given word.

As for sentential semantics, similarly, while logic-based relations between sentences like entailment are more discrete in nature, similarity relations between sentences are clearly on a continuum. LLMs are good at modeling this continuum, as measured in the Semantic Textual Similarity task (STS). One of the tasks in the GLUE benchmark is STS, using data from \cite{cer-etal-2017-semeval}, which consists of pairs of sentences and human-annotated similarity scores. All top 20 models in the leaderboard of GLUE achieve a correlation of 0.91 or more with the human data (both Spearman and Pearson);
\footnote{\url{https://gluebenchmark.com/leaderboard}, retrieved Sept 19 2025. GLUE is described in \cite{wang-etal-2018-glue}.}
human correlation is 0.93. Note that GLUE evaluates models simultaneously on a range of linguistic tasks; these models perform well at STS while at the same time performing well on a range of other natural language tasks, including entailment (NLI, RTE).
This is further evidence for the synthesis view, this time from the point of view of model behavior rather than internal representations and processing.

Turning to non-compositional linguistic processing, a phenomenon that has received a lot of attention in NLP are so-called Multi-Word Expressions~\cite{Villavicencio2005} like \textit{United Arab Emirates}, which have syntactic structure but often function as a single linguistic unit. LLMs perform well at MWE-related tasks like MWE detection~\cite{tayyar-madabushi-etal-2022-semeval}; moreover, \citet{ide-etal-2025-coam} show that a fine-tuned LLM outperforms the previously best system at MWE identification in a varied MWE corpus.
This previous system included a rule-based component and a specifically trained neural network component; again, this suggests that LLMs implicitly implement more symbolic and more distributed processing, and this is benefitial for non-compositional and semi-compositional phenomena. Similarly, LLMs show strong performance at Named Entity Recognition~\cite{malmasi-etal-2022-multiconer}, another well-studied non-compositional phenomenon. Taken together, this suggests that LLMs do non-trivial processing of non-compositional aspects of language, too.

\end{document}